\documentclass[11pt,a4paper]{article}
\usepackage{emnlp2017}
\usepackage{times}
\usepackage{latexsym}

\emnlpfinalcopy

\def\emnlppaperid{704}

\usepackage{lettrine}
\usepackage{algorithm2e}
\usepackage{amsmath}
\usepackage{amsthm}
\usepackage{amsfonts}
\usepackage{amssymb}
\usepackage{mathpartir}
\usepackage{mathtools}
\usepackage{mathrsfs}
\usepackage{mathtools}
\usepackage{siunitx}
\usepackage{pgfplots}

\usepackage[utf8]{inputenc}
\usepackage[T1]{fontenc}

\usepackage{multirow}

\usepackage{color}

\usepackage{xspace}
\usepackage[hang]{footmisc}

\newcommand{\Figure}[1]{Figure~\ref{#1}\xspace}
\newcommand{\Table}[1]{Table~\ref{#1}\xspace}

\title{Regularization techniques for fine-tuning in neural machine translation}

\author{
Antonio Valerio Miceli Barone \quad Barry Haddow \\
\bf Ulrich Germann \quad Rico Sennrich\\
School of Informatics, The University of Edinburgh\\
{\tt \{amiceli, bhaddow, ugermann\}@inf.ed.ac.uk}\\
{\tt rico.sennrich@ed.ac.uk}\\
}

\date{}

\begin{document}
\maketitle
\begin{abstract}
  We investigate techniques for supervised domain adaptation for neural machine translation where an existing model trained on a large out-of-domain dataset is adapted to a small in-domain dataset.
  
In this scenario, overfitting is a major challenge. We investigate a number of techniques to reduce overfitting and improve transfer learning, including regularization techniques such as dropout and L2-regularization towards an out-of-domain prior. In addition, we introduce \emph{tuneout}, a novel regularization technique inspired by dropout.
We apply these techniques, alone and in combination, to neural machine translation, obtaining improvements on IWSLT datasets for English$\rightarrow$German and English$\rightarrow$Russian.
We also investigate the amounts of in-domain training data needed for domain adaptation in NMT, and find a logarithmic relationship between the amount of training data and gain in BLEU score.
\end{abstract}

\section{Introduction}
\label{SEC:INTRO}

Neural machine translation \cite{Bahdanau2014,DBLP:conf/nips/SutskeverVL14} has established itself as the new state of the art at recent shared translation tasks \cite{bojar-EtAl:2016:WMT1,iwslt-report}.
In order to achieve good generalization accuracy, neural machine translation, like most other large machine learning systems, requires large amounts of training examples sampled from a distribution as close as possible to the distribution of the inputs seen during execution.
However, in many applications, only a small amount of parallel text is available for the specific application domain, and it is therefore desirable to leverage larger out-domain datasets.

Owing to the incremental nature of stochastic gradient-based training algorithms, a simple yet effective approach to transfer learning for neural networks is \textit{fine-tuning} \cite{Hinton2006,Mesnil2012,Yosinski2014}: to continue training an existing model which was trained on out-of-domain data with in-domain training data.
This strategy was also found to be very effective for neural machine translation \cite{luong2015,2015arXiv151106709S}.

Since the amount of in-domain data is typically small, overfitting is a concern.
A common solution is early stopping on a small held-out in-domain validation dataset, but this reduces the amount of in-domain data available for training.

In this paper, we show that we can make fine-tuning strategies for neural machine translation more robust by using several regularization techniques.
We consider fine-tuning with varying amounts of in-domain training data, showing that improvements are logarithmic in the amount of in-domain data.

We investigate techniques where domain adaptation starts from a pre-trained out-domain model, and only needs to process the in-domain corpus.
Since we do not need to process the large out-domain corpus during adaptation, this is suitable for scenarios where adaptation must be performed quickly or where the original out-domain corpus is not available.
Other works consider techniques that jointly train on the out-domain and in-domain corpora, distinguishing them using specific input features \cite{DBLP:journals/corr/abs-0907-1815, Finkel:2009:HBD:1620754.1620842, wuebker2015hierarchical}.
These techniques are largely orthogonal to ours\footnote{although in the special case of linear models, they are related to MAP-L2 fine-tuning.} and can be used in combination. In fact,\linebreak[4] \citet{DBLP:journals/corr/ChuDK17} successfully apply fine-tuning in combination with joint training.

\section{Regularization Techniques for Transfer Learning}

Overfitting to the small amount of in-domain training data that may be available is a major challenge in transfer learning for domain adaptation.
We investigate the effect of different regularization techniques to reduce overfitting, and improve the quality of transfer learning.

\subsection{Dropout}
\label{SEC:INTRO:DROPOUT}

The first variant that we consider is fine-tuning with dropout.
Dropout \cite{Srivastava2014} is a stochastic regularization technique for neural networks.
In particular, we consider "Bayesian" dropout for recurrent neural networks \cite{Gal2016}.

In this technique, during training, the columns of the weight matrices of the neural network are randomly set to zero, independently for each example and each epoch, but with the caveat that when the same weight matrix appears multiple times in the unrolled computational graph of a given example, the same columns are zeroed.

For an arbitrary layer that takes an input vector $h$ and computes the pre-activation vector $v$ (ignoring the bias parameter),
\begin{equation}
v_{i,j} = W \cdot M_{W,i,j} \cdot h_{i,j}
\end{equation}
where $M_{W,i,j} = \frac{1}{p} \text{diag}(\text{Bernoulli}^{\otimes n}(p))$ is the dropout mask for matrix $W$ and training example $i$ seen in epoch $j$.
This mask is a diagonal matrix whose entries are drawn from independent Bernoulli random variables with probability $p$ and then scaled by $1/p$.
\newcite{Gal2016} have shown that this corresponds to approximate variational Bayesian inference over the weight matrices considered as model-wise random variables, where the individual weights have a Gaussian prior with zero mean and small diagonal covariance. During execution we simply set the dropout masks to identity matrices, as in the standard approximation scheme.

Since dropout is not a specific transfer learning technique per se, we can apply it during fine-tuning, irrespective of whether or not the original out-of-domain model was also trained with dropout.

\subsection{MAP-L2}
\label{SEC:INTRO:MAPL2}
L2-norm regularization is widely used for machine learning and statistical models. For linear models, it corresponds to imposing a diagonal Gaussian prior with zero mean on the weights.
\newcite{Chelba2006} extended this technique to transfer learning by penalizing the weights of the in-domain model by their L2-distance from the weights of the previously trained out-of-domain model.

For each parameter matrix $W$, the penalty term is
\begin{equation}
L_W = \lambda \cdot \left \| W - \hat{W} \right \|^2_2
\label{EQ:DROPOUT}
\end{equation}
where $W$ is the in-domain parameter matrix to be learned and $\hat{W}$ is the corresponding fixed out-of-domain parameter matrix.
Bias parameters may be regularized as well. For linear models, this corresponds to maximum a posteriori inference w.r.t. a diagonal Gaussian prior with mean equal to the out-of-domain parameters and $1/\lambda$ variance.

To our knowledge this method has not been applied to neural networks, except for a recent work by \newcite{Kirkpatrick2016} which investigates a variant of it for \textit{continual learning} (learning a new task while preserving performance on previously learned task) rather than domain adaptation.
In this work we investigate L2-distance from out-of-domain penalization (MAP-L2) as a domain adaptation technique for neural machine translation.

\subsection{Tuneout}
\label{SEC:INTRO:TUNEOUT}
We also propose a novel transfer learning technique which we call \textit{tuneout}.
Like Bayesian dropout, we randomly drop columns of the weight matrices during training, but instead of setting them to zero, we set them to the corresponding columns of the out-of-domain parameter matrices.

This can be alternatively seen as learning matrices of parameter differences between in-domain and out-of-domain models with standard dropout, starting from a zero initialization at the beginning of fine-tuning.
Therefore, equation \ref{EQ:DROPOUT} becomes
\begin{equation}
v_{i,j} = (\hat{W} + \Delta W \cdot M_{\Delta W,i,j}) \cdot h_{i,j}
\end{equation}
where $\hat{W}$ is the fixed out-of-domain parameter matrix and $\Delta W$ is the parameter difference matrix to be learned and $M_{\Delta W,i,j}$ is a Bayesian dropout mask.

\section{Evaluation}
\label{SEC:EVAL}

We evaluate transfer learning on test sets from the IWSLT shared translation task \cite{cettoloEtAl:EAMT2012}.

\subsection{Data and Methods}
\label{SEC:EVAL:METHODS}

Test sets consist of transcripts of TED talks and their translations; small amounts of in-domain training data are also provided.
For English-to-German we use IWSLT 2015 training data, while for English-to-Russian we use IWSLT 2014 training data.
For the out-of-domain systems, we use training data from the WMT shared translation task,\kern-.25ex\footnote{\url{http://www.statmt.org/wmt16/}} which is considered permissible for IWSLT tasks,
including back-translations of monolingual training data \cite{2015arXiv151106709S}, i.e., automatic translations of data available only in target language ``back'' into the source language.\kern-.25ex\footnote{\url{http://data.statmt.org/rsennrich/wmt16_backtranslations/}}.

We train out-of-domain systems following tools and hyperparameters reported by \newcite{Sennrich2016}, using Nematus \cite{nematus} as the neural machine translation toolkit.
We differ from their setup only in that we use Adam \cite{DBLP:journals/corr/KingmaB14} for optimization.
Our baseline fine-tuning models use the same hyperparameters, except that the learning rate is $4$ times smaller and the validation frequency for early stopping $4$ times higher.
Early stopping serves an important function as the only form of regularization in the baseline fine-tuning model.
We also use this configuration for the in-domain only baselines.

After some exploratory experiments for English-to-German, we set dropout retention probabilities to $0.9$ for word-dropout and $0.8$ for all the other parameter matrices.
Tuneout retention probabilities are set to $0.6$ (word-dropout) and $0.2$ (other parameters).
For MAP-L2 regularization, we found that a penalty of $10^{-3}$ per mini-batch performs best.
For English-to-Russian, retention probabilities of $0.95$ (word-dropout) $0.89$ (other parameters) for both dropout and tuneout performed best.

The out-of-domain training data consists of about $7.92 M$ sentence pairs for English-to-German and $4.06 M$ sentence pairs for English-to-Russian. In-domain training data is about $206 k$ sentence pairs for English-to-German and $181 k$ sentence pairs for English-to-Russian.
Training data is tokenized, truecased and segmented into subword units using byte-pair encoding (BPE) \cite{DBLP:journals/corr/SennrichHB15}.

For replicability and ease of adoption, we include our implementation of dropout and MAP-L2 in the master branch of Nematus. Tuneout regularization is available in a separate code branch of Nematus.\kern-.25ex\footnote{\url{https://github.com/EdinburghNLP/nematus/tree/tuneout-branch}}

\subsection{Results}
\label{SEC:EVAL:RESULTS}

\begin{table*}[t]
\centering
\caption{English-to-German translation BLEU scores}
\label{TABLE:BLEU:ENDE}
\begin{tabular}{lc|ccc|c}
& valid & \multicolumn{4}{c}{test}\\
System & \footnotesize{tst2010} & \footnotesize{tst2011} & \footnotesize{tst2012} & \footnotesize{tst2013} & \footnotesize{avg} \\
\hline
Out-of-domain only & 27.19 & 29.65 & 25.78 & 27.85 & 27.76\\
In-domain only & 25.95 & 27.84 & 23.68 & 25.83 & 25.78\\
Fine-tuning & 30.53 & 32.62 & 28.86 & 32.11 & 31.20\\ 
\hline
Fine-tuning + dropout & 30.63 & 33.06 & 28.90 & 32.02 & 31.33\\ 
Fine-tuning + MAP-L2 & 30.81 & 32.87 & 28.99 & 31.88 & 31.25\\ 
Fine-tuning + tuneout & 30.49 & 32.07 & 28.66 & 31.60 & 30.78$\dagger$\\ 
\hline
Fine-tuning + dropout + MAP-L2 & 30.80 & \textbf{33.19} & \textbf{29.13} & \textbf{32.13} & \textbf{31.48}$\dagger$ \\ 

\hline
\multicolumn{6}{l}{$\dagger$: different from the fine-tuning baseline at $5\%$ significance.}
\end{tabular}
\end{table*}

\begin{table*}[t]
\centering
\caption{English-to-Russian translation BLEU scores}
\begin{tabular}{lc|ccc|c}
& valid & \multicolumn{4}{c}{test}\\
System & \footnotesize{dev2010} & \footnotesize{tst2011} & \footnotesize{tst2012} & \footnotesize{tst2013} & \footnotesize{avg} \\
\hline
Out-of-domain only & 15.74 & 17.48 & 15.15 & 17.81 & 16.81\\
Fine-tuning & 17.47 & 19.67 & 17.17 & 19.18 & 18.67\\ 
\hline
Fine-tuning + dropout & 17.68 & \textbf{19.96} & 17.11 & 19.32 & 18.80\\ 
Fine-tuning + MAP-L2 & \textbf{17.77} & 19.91 & 17.34 & 19.49 & 18.91$\dagger$\\ 
Fine-tuning + tuneout & 17.51 & 19.72 & 17.27 & 19.35 & 18.78\\ 
\hline
Fine-tuning + dropout + MAP-L2 & 17.74 & 19.68 & \textbf{17.83} & \textbf{19.78} & \textbf{19.10}$\dagger$\\ 
\hline
\multicolumn{6}{l}{ $\dagger$: different from the fine-tuning baseline at $5\%$ significance.}
\end{tabular}
\label{TABLE:BLEU:ENRU}
\end{table*}

We report the translation quality in terms of NIST-BLEU scores of our models in \Table{TABLE:BLEU:ENDE} for English-to-German and \Table{TABLE:BLEU:ENRU} for English-to-Russian.
Statistical significance on the concatenated test sets scores is determined via bootstrap resampling \cite{koehn2004statistical}.

Dropout and MAP-L2 improve translation quality when fine-tuning both separately and in combination.
When the two methods are used in combination, the improvements are significant at $5\%$ for both language pairs, while in isolation dropout is non-significant and MAP-L2 is only significant for English-to-Russian.
Tuneout does not yield improvements for English-to-German, in fact it is significantly worse, but yields a small, non-significant improvement for English-to-Russian.

\begin{figure}
  
\begin{tikzpicture}[scale=0.9]
\pgfplotsset{major grid style={style=dotted,color=black!20}}
\begin{axis}[xlabel=Training mini-batches ($\cdot 10^4$),
    ymin = 26.5,
    ymax = 31.5,
    xmin = 0,
    xmax = 11.3,
    ylabel={\sc Bleu},
    legend pos = south east,
    legend style={
        font=\scriptsize,
        /tikz/nodes={anchor=west}
        },
    mark size = 0.1,
    ]

    \addplot +[black, mark size=1pt, mark color=black, mark=*, solid, line width=0.2ex] table[x expr=\thisrow{Minibatch}/10000] {ende-bleus-baseline.txt};

    \addplot +[red, mark size=2pt, mark options={solid}, mark color=black, mark=|, line width=0.15ex] table[x expr=\thisrow{Minibatch}/10000] {ende-bleus-dropout.txt};
    
    \addplot +[blue, mark size=2pt, mark options={solid}, mark color=black, mark=x, line width=0.15ex] table[x expr=\thisrow{Minibatch}/10000] {ende-bleus-map.txt};
    
    \addplot +[green, mark size=1pt, mark options={solid}, mark color=black, mark=triangle*, line width=0.15ex] table[x expr=\thisrow{Minibatch}/10000] {ende-bleus-tuneout-veryhigh.txt};

    \addlegendentry{fine-tune}
    \addlegendentry{fine-tune + dropout}
    \addlegendentry{fine-tune + L2-MAP}
    \addlegendentry{fine-tune + tuneout}

\end{axis}
\end{tikzpicture}
\caption{English$\rightarrow$German validation {\sc Bleu} over training mini-batches.}
\label{FIG:VALIDBLEU:ENDE}
\end{figure}
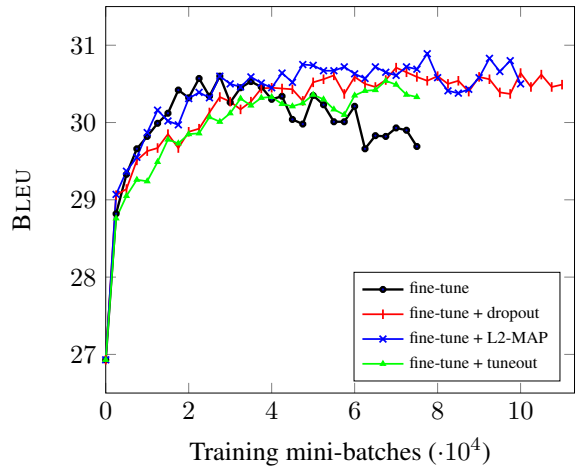

In order to obtain a better picture  of the training dynamics, we plot training curves\footnote{These BLEU scores are computed using Moses \texttt{multi-bleu.perl} which gives slightly different results than NIST \texttt{mteval-v13a.pl} that is used for \Table{TABLE:BLEU:ENDE}.}
for several of our English-to-German models in \Figure{FIG:VALIDBLEU:ENDE}.
Baseline fine-tuning starts to noticeably overfit between the second and third epoch ($1$ epoch $\approx 10^4$ mini-batches), while dropout, MAP-L2 and tuneout seem to converge without displaying noticeable overfitting.

In our experiments, all forms of regularization, including early stopping, have shown to be successful at mitigating the effect of overfitting.
Still, our results suggest that there is value in not relying only on early stopping:

\begin{itemize}
\item our results suggest that multiple regularizers outperform a single one.
\item if the amount of in-domain data is very small, we may want to use all of it for fine-tuning, and not hold out any for early stopping.
\end{itemize}

\begin{figure}
\vspace*{-1em}

\begin{tikzpicture}[scale=0.9]
\pgfplotsset{major grid style={style=dotted,color=black!20}}
\begin{semilogxaxis}[xlabel=In-domain training sentence pairs,
    ymin = 25.0,
    ymax = 30.5,
    xmin = 10,
    xmax = 205678,
    ylabel={\sc Bleu},
    legend pos = north west,
    legend style={
        font=\scriptsize,
        /tikz/nodes={anchor=west}
        },
    mark size = 0.1,
    ]

    \draw[black] (axis cs:0,25.27) -- (axis cs:205678,25.27) node [above left] {\scriptsize baseline};

    \addplot +[black, mark size=1pt, mark options={solid}, mark color=black, mark=*, dotted, line cap=round, line width=0.2ex] table[x expr=\thisrow{Sentence-pairs}] {EMNLP_2017_datapoints.noreg.epoch-1.txt};
    
    \addplot +[black, mark size=1pt, mark options={solid}, mark color=black, mark=triangle*, dashed, line width=0.2ex] table[x expr=\thisrow{Sentence-pairs}] {EMNLP_2017_datapoints.noreg.epoch-5.txt};
    
    \addplot +[black, mark size=2pt, mark options={solid}, mark color=black, mark=|, line width=0.15ex] table[x expr=\thisrow{Sentence-pairs}] {EMNLP_2017_datapoints.noreg.txt};
    
    \addplot +[red, mark size=2pt, mark options={solid}, mark color=black, mark=x, line width=0.15ex] table[x expr=\thisrow{Sentence-pairs}] {EMNLP_2017_datapoints.reg.txt};
    
    \addplot +[brown, mark size=2pt, mark options={solid}, mark color=black, mark=|, dashed, line width=0.2ex] table[x expr=\thisrow{Sentence-pairs}] {EMNLP_2017_datapoints.reg.epoch-5.txt};

    \addlegendentry{1 epoch}
    \addlegendentry{5 epochs}
    \addlegendentry{early stop}
    \addlegendentry{early stop + dropout + L2-MAP}
    \addlegendentry{5 epochs  + dropout + L2-MAP}

\end{semilogxaxis}
\end{tikzpicture}
\caption{English$\rightarrow$German test {\sc Bleu} with fine-tuning on different in-domain data set size. Baseline trained on WMT data.}
\label{FIG:SPVSBLEUTEST:ENDE}
\vspace*{-1em}
\end{figure}
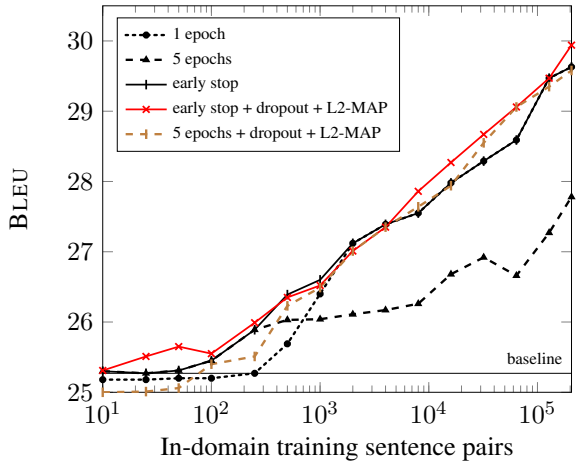

\noindent
To evaluate different fine-tuning streategies on varying amounts of in-domain data, we tested fine-tuning with random samples of in-domain data, ranging from 10 sentence pairs to the full data set of $206 k$ sentence pairs.
Fine-tuning with low amounts of training data is of special interest for online adaptation scenarios where a system is fed back post-edited translation.\footnote{We expect even bigger gains in that scenario because we would not train on a random sample, but on translations that are conceivably from the same document.}
Results are shown in \Figure{FIG:SPVSBLEUTEST:ENDE}.

The results show an approximately logarithmic relation between the size of the in-domain training set and BLEU.
We consider three baseline approaches: fine-tuning for a fixed number of epochs (1 or 5), or early stopping.
All three baseline approaches have their disadvantages.
Fine-tuning for 1 epoch shows underfitting on small amounts of data (less than 1,000 sentence pairs); fine-tuning for 5 epochs overfits on 500-200,000 sentence pairs.
Early stopping is generally a good strategy, but it requires an in-domain held-out dataset.

On the same amount of data, regularization (dropout+MAP-L2) leads to performance that is better (or no worse) than the baseline with only early stopping.
Fine-tuning with regularization is also more stable, and if we have no access to a in-domain valdiation set for early stopping, can be run for a fixed number of epochs with little or no accuracy loss.

\section{Conclusion}
\label{SEC:CONCLUSION}
We investigated fine-tuning for domain adaptation in neural machine translation with different amounts of in-domain training data, and strategies to avoid overfitting.
We found that our baseline that relies only on early stopping has a strong performance, but fine-tuning with recurrent dropout and with MAP-L2 regularization yield additional small improvements of the order of $0.3$ BLEU points for both English-to-German and English-to-Russian, while the improvements in terms of final translation accuracy of tuneout appear to be less consistent.

Furthermore, we found that regularization techniques that we considered make training more robust to overfitting, which is particularly helpful in scenarios where only small amounts of in-domain data is available, making early-stopping impractical as it relies on a sufficiently large in-domain validation set.
Given the results of our experiments, we recommend using both dropout and MAP-L2 regularization for fine-tuning tasks, since they are easy to implement, efficient, and yield improvements while stabilizing training.
We also present a learning curve that shows a logarithmic relationship between the amount of in-domain training data and the quality of the adapted system.

Our techniques are not specific to neural machine translation, and we propose that they could be also tried for other neural network architectures and other tasks.

\section*{Acknowledgments}
\lettrine[image=true, lines=2, findent=1ex, nindent=0ex, loversize=.15]{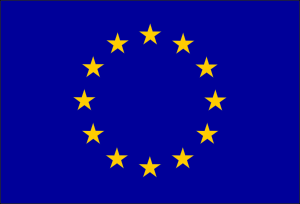}%
{T}his project has received funding from the European Union's Horizon
2020 research and innovation programme under grant agreements 644333 (TraMOOC) and 645487 (ModernMT).
We also thank Booking.com for their support.

\bibliography{tuneout}

\begin{thebibliography}{23}
\expandafter\ifx\csname natexlab\endcsname\relax\def\natexlab#1{#1}\fi

\bibitem[{Bahdanau et~al.(2015)Bahdanau, Cho, and Bengio}]{Bahdanau2014}
Dzmitry Bahdanau, Kyunghyun Cho, and Yoshua Bengio. 2015.
\newblock {Neural Machine Translation by Jointly Learning to Align and
  Translate}.
\newblock In \emph{{Proceedings of the International Conference on Learning
  Representations (ICLR)}}.

\bibitem[{Bojar et~al.(2016)Bojar, Chatterjee, Federmann, Graham, Haddow, Huck,
  {Jimeno Yepes}, Koehn, Logacheva, Monz, Negri, Neveol, Neves, Popel, Post,
  Rubino, Scarton, Specia, Turchi, Verspoor, and
  Zampieri}]{bojar-EtAl:2016:WMT1}
Ond\v{r}ej Bojar, Rajen Chatterjee, Christian Federmann, Yvette Graham, Barry
  Haddow, Matthias Huck, Antonio {Jimeno Yepes}, Philipp Koehn, Varvara
  Logacheva, Christof Monz, Matteo Negri, Aurelie Neveol, Mariana Neves, Martin
  Popel, Matt Post, Raphael Rubino, Carolina Scarton, Lucia Specia, Marco
  Turchi, Karin Verspoor, and Marcos Zampieri. 2016.
\newblock {Findings of the 2016 Conference on Machine Translation (WMT16)}.
\newblock In \emph{{Proceedings of the First Conference on Machine Translation,
  Volume 2: Shared Task Papers}}, pages 131--198, Berlin, Germany. Association
  for Computational Linguistics.

\bibitem[{Cettolo et~al.(2012)Cettolo, Girardi, and
  Federico}]{cettoloEtAl:EAMT2012}
Mauro Cettolo, Christian Girardi, and Marcello Federico. 2012.
\newblock {WIT$^3$: Web Inventory of Transcribed and Translated Talks}.
\newblock In \emph{{Proceedings of the 16$^{th}$ Conference of the European
  Association for Machine Translation (EAMT)}}, pages 261--268, Trento, Italy.

\bibitem[{Cettolo et~al.(2016)Cettolo, Niehues, St{\"u}ker, Bentivogli, and
  Federico}]{iwslt-report}
Mauro Cettolo, Jan Niehues, Sebastian St{\"u}ker, Luisa Bentivogli, and
  Marcello Federico. 2016.
\newblock {Report on the 13th IWSLT Evaluation Campaign}.
\newblock In \emph{{IWSLT 2016}}, Seattle, USA.

\bibitem[{Chelba and Acero(2006)}]{Chelba2006}
Ciprian Chelba and Alex Acero. 2006.
\newblock {Adaptation of maximum entropy capitalizer: Little data can help a
  lot}.
\newblock \emph{Computer Speech \& Language}, 20(4):382--399.

\bibitem[{Chu et~al.(2017)Chu, Dabre, and
  Kurohashi}]{DBLP:journals/corr/ChuDK17}
Chenhui Chu, Raj Dabre, and Sadao Kurohashi. 2017.
\newblock {An Empirical Comparison of Simple Domain Adaptation Methods for
  Neural Machine Translation}.
\newblock In \emph{{Proceedings of the 54th Annual Meeting of the Association
  for Computational Linguistics}}, Vancouver, Canada.

\bibitem[{{Daume III}(2007)}]{DBLP:journals/corr/abs-0907-1815}
Hal {Daume III}. 2007.
\newblock {Frustratingly Easy Domain Adaptation}.
\newblock In \emph{{Proceedings of the 45th Annual Meeting of the Association
  of Computational Linguistics}}, pages 256--263, Prague, Czech Republic.
  Association for Computational Linguistics.

\bibitem[{Finkel and Manning(2009)}]{Finkel:2009:HBD:1620754.1620842}
Jenny~Rose Finkel and Christopher~D. Manning. 2009.
\newblock {Hierarchical Bayesian Domain Adaptation}.
\newblock In \emph{{Proceedings of Human Language Technologies: The 2009 Annual
  Conference of the North American Chapter of the Association for Computational
  Linguistics}}, {NAACL '09}, pages 602--610, Stroudsburg, PA, USA. Association
  for Computational Linguistics.

\bibitem[{Gal and Ghahramani(2016)}]{Gal2016}
Yarin Gal and Zoubin Ghahramani. 2016.
\newblock {A Theoretically Grounded Application of Dropout in Recurrent Neural
  Networks}.
\newblock In \emph{{Advances in Neural Information Processing Systems 29
  (NIPS)}}.

\bibitem[{Hinton and Salakhutdinov(2006)}]{Hinton2006}
Geoffrey~E Hinton and Ruslan~R Salakhutdinov. 2006.
\newblock {Reducing the dimensionality of data with neural networks}.
\newblock \emph{Science}, 313(5786):504--507.

\bibitem[{Kingma and Ba(2015)}]{DBLP:journals/corr/KingmaB14}
Diederik~P. Kingma and Jimmy Ba. 2015.
\newblock {Adam: {A} Method for Stochastic Optimization}.
\newblock In \emph{{The International Conference on Learning Representations}},
  San Diego, California, USA.

\bibitem[{Kirkpatrick et~al.(2017)Kirkpatrick, Pascanu, Rabinowitz, Veness,
  Desjardins, Rusu, Milan, Quan, Ramalho, Grabska{-}Barwinska, Hassabis,
  Clopath, Kumaran, and Hadsell}]{Kirkpatrick2016}
James Kirkpatrick, Razvan Pascanu, Neil~C. Rabinowitz, Joel Veness, Guillaume
  Desjardins, Andrei~A. Rusu, Kieran Milan, John Quan, Tiago Ramalho, Agnieszka
  Grabska{-}Barwinska, Demis Hassabis, Claudia Clopath, Dharshan Kumaran, and
  Raia Hadsell. 2017.
\newblock {Overcoming catastrophic forgetting in neural networks}.
\newblock \emph{Proceedings of the National Academy of Sciences},
  114(13):3521--3526.

\bibitem[{Koehn(2004)}]{koehn2004statistical}
Philipp Koehn. 2004.
\newblock {Statistical Significance Tests for Machine Translation Evaluation.}
\newblock In \emph{{Proceedings of EMNLP 2004}}, pages 388--395, Barcelona,
  Spain.

\bibitem[{Luong and Manning(2015)}]{luong2015}
Minh-Thang Luong and Christopher~D. Manning. 2015.
\newblock {Stanford Neural Machine Translation Systems for Spoken Language
  Domains}.
\newblock In \emph{{Proceedings of the International Workshop on Spoken
  Language Translation 2015}}, Da Nang, Vietnam.

\bibitem[{Mesnil et~al.(2012)Mesnil, Dauphin, Glorot, Rifai, Bengio,
  Goodfellow, Lavoie, Muller, Desjardins, Warde-Farley et~al.}]{Mesnil2012}
Gr{\'e}goire Mesnil, Yann Dauphin, Xavier Glorot, Salah Rifai, Yoshua Bengio,
  Ian~J Goodfellow, Erick Lavoie, Xavier Muller, Guillaume Desjardins, David
  Warde-Farley, et~al. 2012.
\newblock {Unsupervised and Transfer Learning Challenge: a Deep Learning
  Approach.}
\newblock \emph{ICML Unsupervised and Transfer Learning}, 27:97--110.

\bibitem[{Sennrich et~al.(2017)Sennrich, Firat, Cho, Birch, Haddow, Hitschler,
  Junczys-Dowmunt, L{\"a}ubli, {Miceli Barone}, Mokry, and Nadejde}]{nematus}
Rico Sennrich, Orhan Firat, Kyunghyun Cho, Alexandra Birch, Barry Haddow,
  Julian Hitschler, Marcin Junczys-Dowmunt, Samuel L{\"a}ubli, Antonio~Valerio
  {Miceli Barone}, Jozef Mokry, and Maria Nadejde. 2017.
\newblock {Nematus: a Toolkit for Neural Machine Translation}.
\newblock In \emph{{Proceedings of the Software Demonstrations of the 15th
  Conference of the European Chapter of the Association for Computational
  Linguistics}}, pages 65--68, Valencia, Spain.

\bibitem[{Sennrich et~al.(2016{\natexlab{a}})Sennrich, Haddow, and
  Birch}]{Sennrich2016}
Rico Sennrich, Barry Haddow, and Alexandra Birch. 2016{\natexlab{a}}.
\newblock {Edinburgh Neural Machine Translation Systems for WMT 16}.
\newblock In \emph{{Proceedings of the First Conference on Machine
  Translation}}, pages 371--376, Berlin, Germany. Association for Computational
  Linguistics.

\bibitem[{Sennrich et~al.(2016{\natexlab{b}})Sennrich, Haddow, and
  Birch}]{2015arXiv151106709S}
Rico Sennrich, Barry Haddow, and Alexandra Birch. 2016{\natexlab{b}}.
\newblock {Improving Neural Machine Translation Models with Monolingual Data}.
\newblock In \emph{{Proceedings of the 54th Annual Meeting of the Association
  for Computational Linguistics (Volume 1: Long Papers)}}, pages 86--96,
  Berlin, Germany. Association for Computational Linguistics.

\bibitem[{Sennrich et~al.(2016{\natexlab{c}})Sennrich, Haddow, and
  Birch}]{DBLP:journals/corr/SennrichHB15}
Rico Sennrich, Barry Haddow, and Alexandra Birch. 2016{\natexlab{c}}.
\newblock {Neural Machine Translation of Rare Words with Subword Units}.
\newblock In \emph{{Proceedings of the 54th Annual Meeting of the Association
  for Computational Linguistics (Volume 1: Long Papers)}}, pages 1715--1725,
  Berlin, Germany. Association for Computational Linguistics.

\bibitem[{Srivastava et~al.(2014)Srivastava, Hinton, Krizhevsky, Sutskever, and
  Salakhutdinov}]{Srivastava2014}
Nitish Srivastava, Geoffrey~E Hinton, Alex Krizhevsky, Ilya Sutskever, and
  Ruslan Salakhutdinov. 2014.
\newblock {Dropout: a simple way to prevent neural networks from overfitting.}
\newblock \emph{Journal of Machine Learning Research}, 15(1):1929--1958.

\bibitem[{Sutskever et~al.(2014)Sutskever, Vinyals, and
  Le}]{DBLP:conf/nips/SutskeverVL14}
Ilya Sutskever, Oriol Vinyals, and Quoc~V. Le. 2014.
\newblock {Sequence to Sequence Learning with Neural Networks}.
\newblock In \emph{{Advances in Neural Information Processing Systems 27:
  Annual Conference on Neural Information Processing Systems 2014}}, pages
  3104--3112, Montreal, Quebec, Canada.

\bibitem[{Wuebker et~al.(2015)Wuebker, Green, and
  DeNero}]{wuebker2015hierarchical}
Joern Wuebker, Spence Green, and John DeNero. 2015.
\newblock {Hierarchical Incremental Adaptation for Statistical Machine
  Translation.}
\newblock In \emph{{Proceedings of the 2015 Conference on Empirical Methods in
  Natural Language Processing}}, pages 1059--1065, Lisbon, Portugal.

\bibitem[{Yosinski et~al.(2014)Yosinski, Clune, Bengio, and
  Lipson}]{Yosinski2014}
Jason Yosinski, Jeff Clune, Yoshua Bengio, and Hod Lipson. 2014.
\newblock {How transferable are features in deep neural networks?}
\newblock In \emph{{Advances in neural information processing systems}}, pages
  3320--3328.

\end{thebibliography}
\bibliographystyle{ugHarv3plainEtal}

\end{document}